\documentclass[11pt]{article}
\usepackage[final]{acl}
\usepackage{times}
\usepackage{latexsym}
\usepackage[T1]{fontenc}
\usepackage[utf8]{inputenc}
\usepackage{microtype}
\usepackage{inconsolata}
\usepackage{graphicx}
\usepackage{booktabs}
\usepackage{multirow}
\usepackage{amsmath}
\usepackage{amssymb}
\usepackage{enumitem}
\usepackage{url}

\newcommand{\mccxr}{\textsc{MC-CXR}}
\newcommand{\micxr}{\textsc{MI-CXR}}

\title{MC-CXR: A Multi-Context Chest X-ray Benchmark for Context-Induced Disruption in Vision--Language Models}

\author{
  Junhyeok Lee$^{1,2*}$ \quad Songsoo Kim$^{2*}$ \quad Kyu Sung Choi$^{2,3,4\dagger}$ \\
  $^{1}$Interdisciplinary Program in Cancer Biology, Seoul National University College of Medicine \\
  $^{2}$Department of Radiology, Seoul National University Hospital \\
  $^{3}$Department of Radiology, Seoul National University College of Medicine \\
  $^{4}$Healthcare AI Research Institute, Seoul National University Hospital \\
  Republic of Korea \\
  $^{*}$Equal contribution \quad $^{\dagger}$Corresponding author \\
  \texttt{jhlee0619@snu.ac.kr} \quad \texttt{crown7699@gmail.com} \quad \texttt{ent1127@snu.ac.kr} 
}

\begin{document}
\maketitle

\begin{abstract}
Vision--language models (VLMs) are increasingly used in clinical pipelines where a chest X-ray is interpreted alongside retrieved reports, preliminary notes, or prior imaging. Existing benchmarks measure whether models answer correctly in isolation, but not whether they preserve a correct image-only decision when plausible context conflicts with the image. We introduce Multi-Context Chest X-ray (\mccxr{}), a benchmark of 240 cases expanded into 2{,}522 instances that isolates context-induced disruption through paired perturbation. Each case fixes the current image and target finding while presenting matched reliable and misleading context across text and prior CXR, with visual overlays where available. \mccxr{} defines three task families and two paired metrics, the switch-to-wrong rate and the context-aligned error rate. We evaluate ten VLMs spanning open-source general, medical-domain, and closed-source systems. Image-only accuracy is necessary but insufficient. Mean switch rates range from 45.6--78.1\% across misleading textual sources and 35.7--61.7\% across misleading visual sources. Among switched predictions, 74.6\% align with the misleading label for text versus 17.6\% for visual context, a 57.0-point gap (95\% CI 50.9--62.8). This text--visual asymmetry is observed under the standardized direct-answer protocol. The dataset is available on \href{https://physionet.org/}{PhysioNet}.
\end{abstract}

\section{Introduction}

Chest X-ray (CXR) interpretation is rarely an image-only task. As part of standard clinical practice, radiologists interpret the current image together with the clinical indication, prior imaging, and prior reports \citep{castillo2021clinicalinfo,small2021clinicalhistory} to improve diagnostic accuracy and clinical relevance. This integration is part of the radiologic standard of care rather than an optional aid \citep{test2013clinicalhistory,leslie2000clinicalinfo}, so any system interpreting CXRs within this workflow inherits the same expectation. Modern automated vision--language model (VLM) pipelines increasingly mirror this process, often expanding the context to include preliminary draft notes alongside the image \citep{chen2024chexagent,lee2025cxrllava,singhal2023clinical,singhal2025medpalm}.

\begin{figure}[t]
\centering
\includegraphics[width=\linewidth]{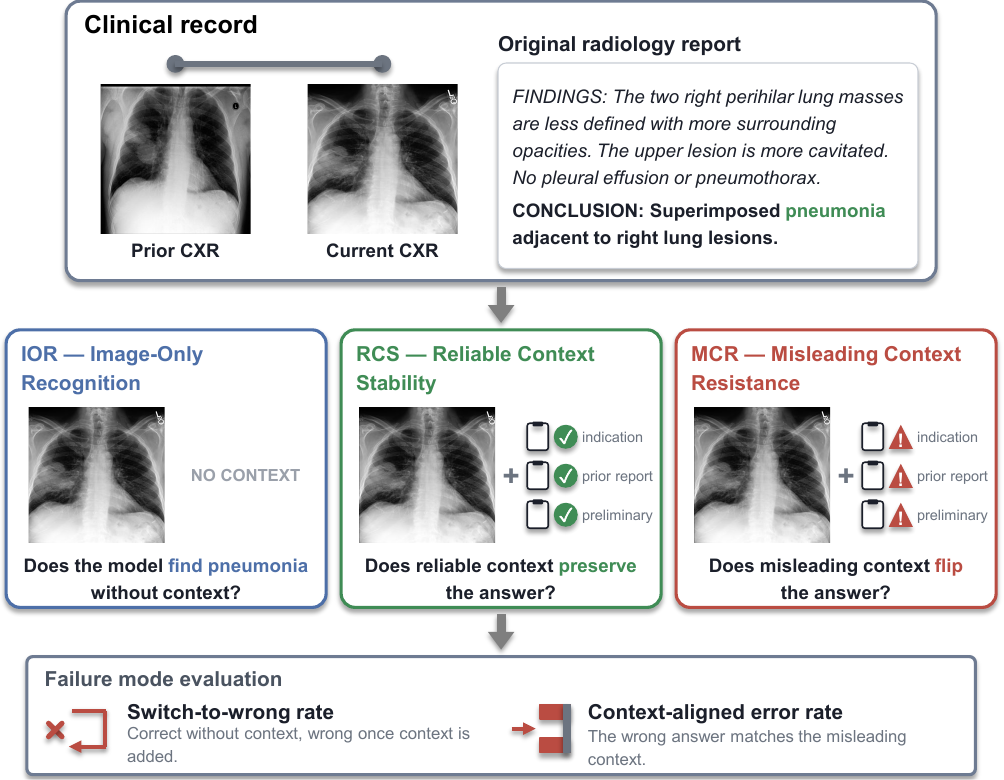}
\caption{Overview of context-robust CXR evaluation and \mccxr{}. Clinical image interpretation often combines current-image evidence with auxiliary context. We formalize context-robust visual reasoning into three task families, Image-only Recognition (IOR), Reliable Context Stability (RCS), and Misleading Context Resistance (MCR), and evaluate them across textual, prior-image, and visual-overlay perturbations.}
\label{fig:overview}
\end{figure}

The challenge is not that clinical context is undesirable, but that its reliability varies. In clinical-style or retrieval-augmented VLM workflows, context can be accurate and useful, but it can also be stale, mismatched to the patient, summarized inaccurately, generated as a preliminary draft, or over-weighted relative to the current image. While these issues collectively cause \emph{context-induced disruption}, in practice we focus on mismatched or misleading auxiliary evidence. In this setting, clinical context pulls a VLM away from image evidence even when the image-only decision is correct. This concern is related to text-only failures studied as sycophancy, where models follow external statements that should not control the answer \citep{sharma2024sycophancy}, and knowledge conflict, where retrieved or in-context evidence overrides a model's own correct prior \citep{xie2024chameleon,wang2025conflictingrag,longpre2021entityconflict}.

Existing CXR benchmarks evaluate recognition \citep{wang2017chestxray8,irvin2019chexpert}, single-image VQA \citep{lau2018vqarad,liu2025gemex,pal2025rexvqa}, report generation \citep{wang2018tienet,chen2020r2gen}, or longitudinal progression \citep{bannur2023mscxrt,cho2026micxr,moon2026lunguage}, but they do not directly test whether a model preserves a correct current-image decision when context is reliable or resists plausible but misleading context when it conflicts with the image. Using aggregate accuracy on these benchmarks as a proxy for context robustness is therefore misleading \citep{degrave2021ai,geirhos2020shortcut,ribeiro2020checklist}. Because such context is increasingly attached automatically, whether retrieved or generated, a model that silently defers to unreliable evidence poses a clinical safety risk that aggregate accuracy cannot reveal.

We call the ability to use context conditionally \emph{context-robust visual reasoning}. It requires four sub-capabilities that ordinary image-only accuracy cannot separate. The model must recognize current-image evidence, parse the context source, identify agreement or disagreement between image and context, and decide which evidence source should determine the answer under conflict. Image-only benchmarks primarily measure the first capability, leaving context parsing, conflict detection, and arbitration untested. Arbitration determines whether the image or the context prevails when the two disagree, and it is what \mccxr{} is designed to measure directly.

To evaluate this capability, we introduce the Multi-Context Chest X-ray benchmark \mccxr{}. Each case fixes a current CXR image $I_i$ and a target finding $X_i$ while systematically varying the reliability and modality of auxiliary context. By pairing reliable and misleading variants of the same context source, \mccxr{} tests conditional context use rather than penalizing context use itself. This paired-perturbation construction extends contrast-set methodology from NLP \citep{ribeiro2020checklist,gardner2020contrastsets,kaushik2020counterfactual} to multimodal medical reasoning \citep{saporta2022saliency}. Figure~\ref{fig:overview} illustrates the benchmark design and the three task families.

Our contributions are as follows:

\begin{itemize}[leftmargin=*]
    \item \textbf{Evidence arbitration as an evaluation target.} We frame context robustness as deciding when reliable context should support current-image evidence and when misleading context should be resisted.
    \item \textbf{A paired-perturbation CXR benchmark.} \mccxr{} contains 240 radiologist-curated cases expanded into 2{,}522 instances, pairing reliable and misleading clinical indications, prior reports, preliminary notes, prior CXRs, and visual overlays.
    \item \textbf{Switch metrics and text--visual asymmetry.} Switch-to-wrong and context-aligned error rates isolate disruption on image-only-correct cases and show that misleading text induces stronger, more label-aligned switching than misleading visual context across ten VLMs.
\end{itemize}

\begin{table*}[t]
\centering
\footnotesize
\setlength{\tabcolsep}{4pt}
\renewcommand{\arraystretch}{1.05}
\begin{tabular*}{\textwidth}{@{\extracolsep{\fill}} l p{0.22\textwidth} c c c c c c @{}}
\toprule
Benchmark & Primary task & Disrupt. & Ind. & Img. & Rep. & Note & Overlay \\
\midrule
ReXVQA \citep{pal2025rexvqa}                    & Multi-task CXR VQA        & -- & -- & -- & -- & -- & -- \\
MS-CXR-T \citep{bannur2023mscxrt}               & Temporal progression      & -- & -- & \checkmark & \checkmark & -- & -- \\
\micxr{} \citep{cho2026micxr}                   & Longitudinal reasoning    & -- & -- & \checkmark & \checkmark & -- & -- \\
\midrule
\textbf{\mccxr{} (ours)}                        & Context-robust reasoning  & \checkmark & \checkmark & \checkmark & \checkmark & \checkmark & \checkmark \\
\bottomrule
\end{tabular*}
\caption{Comparison of \mccxr{} with existing medical VQA and longitudinal CXR benchmarks. Columns indicate whether each benchmark evaluates paired context-induced disruption and which auxiliary inputs it supports across indication, prior image, prior report, preliminary note, and overlay. Among the benchmarks compared here, \mccxr{} combines all five context sources with explicit disruption metrics.}
\label{tab:benchmark_compare}
\end{table*}

\section{Related Work}

\subsection{Medical VQA and CXR Datasets}
\label{sec:related_vqa}

Existing medical VQA and CXR benchmarks evaluate three capabilities in isolation, including image recognition over large labeled cohorts \citep{johnson2019mimiccxr,wang2017chestxray8,irvin2019chexpert}, single-context VQA on CXR and adjacent domains \citep{lau2018vqarad,liu2021slake,liu2025gemex,pal2025rexvqa,zhang2023pmcvqa}, and report generation from a single current image \citep{wang2018tienet,chen2020r2gen}. Across all three axes, the model is given exactly one context source and graded on whether its output matches a label. This protocol cannot distinguish a model that grounds its answer in the image from one that is led to the same answer by context. Aggregate accuracy hides this confound by construction.

\subsection{Contextual and Longitudinal Reasoning in CXR}

Clinical information has long been recognized as part of radiologic interpretation. Systematic reviews and empirical studies show that clinical history shapes reporting accuracy, confidence, relevance, and conclusions, and that inaccurate or absent history can compromise interpretation \citep{castillo2021clinicalinfo,small2021clinicalhistory,test2013clinicalhistory,leslie2000clinicalinfo}. Longitudinal benchmarks evaluate how models use prior studies and reports over time. MS-CXR-T probes temporal structure in biomedical VLMs \citep{bannur2023mscxrt}, Chest ImaGenome offers anatomy-centered scene graphs with chronological relations \citep{wu2021chestimagenome}, and recent benchmarks such as \micxr{} and LUNGUAGE emphasize longitudinal and sequential interpretation \citep{cho2026micxr,moon2026lunguage}. These benchmarks share a common premise. Prior context, whether temporal, structured, or sequential, is assumed reliable, and the evaluation measures how well models exploit it. \mccxr{} treats this premise itself as the variable. By contrasting reliable context conditions with clinically plausible but misleading counterparts, \mccxr{} measures arbitration rather than exploitation. The benchmark tests whether a model tells when context should and should not override its image-only decision.

\subsection{Context Robustness and Evidence Conflict in VLMs}

The NLP literature has long shown that aggregate accuracy hides capability-specific failures through behavioral CheckLists, contrast sets, and shortcut-learning audits \citep{ribeiro2020checklist,gardner2020contrastsets,geirhos2020shortcut}. \mccxr{} adapts these case-internal contrast protocols to the multimodal--clinical setting, where the perturbation axis is context reliability across modalities rather than surface-form text. In text-only LLMs, retrieved or in-context evidence systematically pulls models from their correct priors. This behavior is studied under the labels knowledge conflict \citep{lewis2020rag,yoran2024robustrag,wang2025conflictingrag,zhang2025faithfulrag} and sycophancy \citep{sharma2024sycophancy}. Whether this failure mode crosses into vision--language models is an open empirical question. \mccxr{} addresses it directly. By holding the current image fixed and pairing five clinically realistic context sources across reliable and misleading conditions, it measures context arbitration and reveals stronger directional pull from textual than visual context under the direct-answer protocol. We now translate this gap into a benchmark design that holds the current image fixed while varying the reliability and modality of auxiliary context. Table~\ref{tab:benchmark_compare} summarizes the comparison with existing benchmarks along these dimensions.

\section{MC-CXR Benchmark}
\label{sec:mccxr}

\subsection{Task Formulation}
\label{sec:task_description}

\begin{figure*}[t]
\centering
\includegraphics[width=\linewidth]{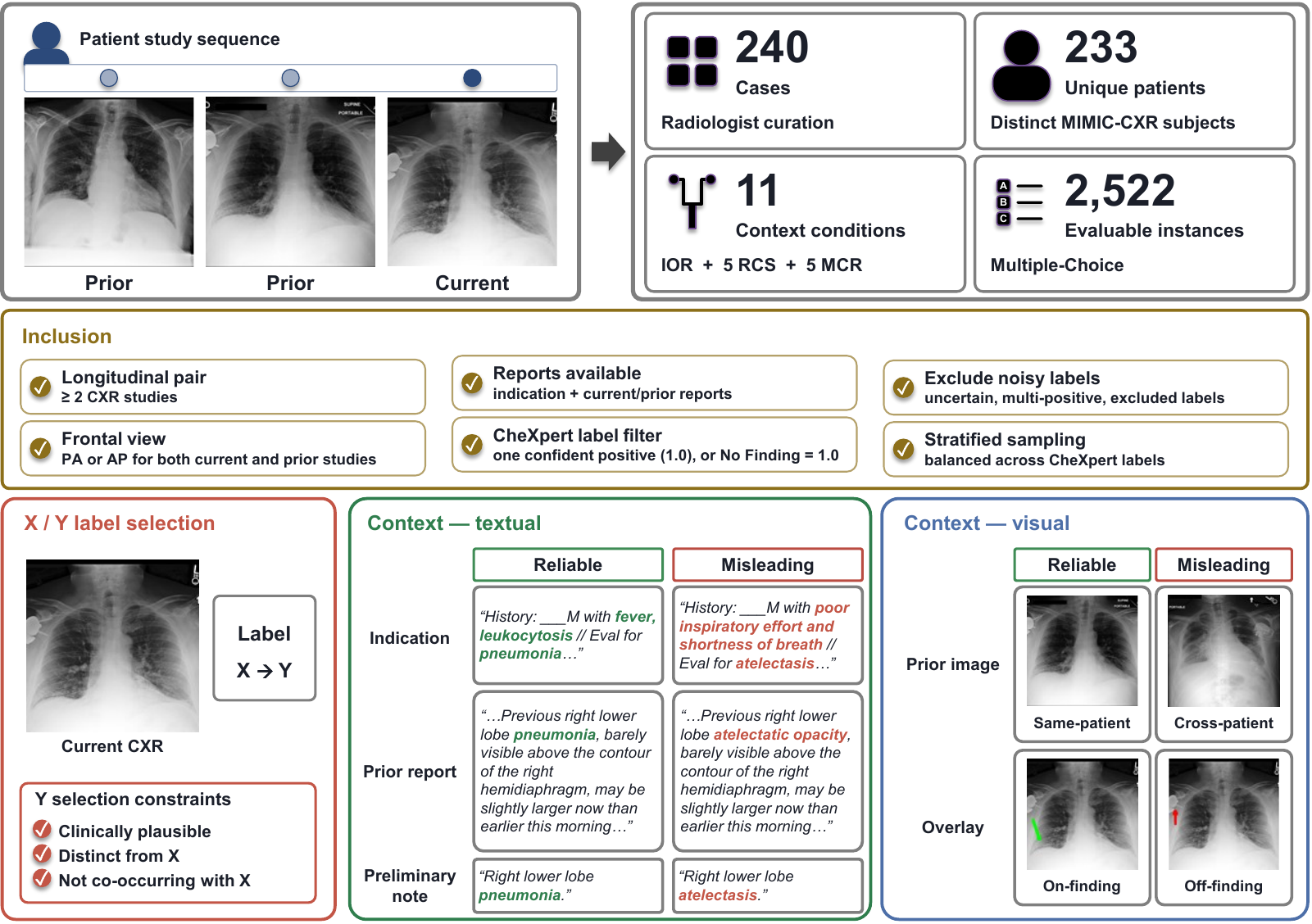}
\caption{Overview of \mccxr{} construction. We derive eligible cases from MIMIC-CXR and MIMIC-CXR-JPG, select target and misleading labels through radiologist review, and instantiate matched reliable and misleading context conditions. The resulting cohort supports paired evaluation of reliable-context stability and misleading-context resistance.}
\label{fig:construction}
\end{figure*}

\mccxr{} evaluates whether a vision--language model preserves a correct current-image decision when additional evidence is provided. Each benchmark case contains a current CXR image $I_i$ from MIMIC-CXR, a target current-image finding $X_i$, and a context-implied finding $Y_i$. Both $X_i$ and $Y_i$ are drawn from the same curated CheXpert-derived label space \citep{irvin2019chexpert}. Context means auxiliary evidence beyond the current unannotated image, and it spans three modalities. Textual information includes the clinical indication, the prior report, and the preliminary note. Visual prior evidence is a same- or different-patient prior CXR. Overlaid visual cues are PACS-style annotations on the current image. Under context condition $c$, the model emits a single predicted label $\hat{p}_{i,c}$ from a 12-label CheXpert subset (prompts in Appendix~\ref{app:templates}). The response is target-correct when $\hat{p}_{i,c}=X_i$. Three task families share this output schema, an image-only baseline plus two robustness axes RCS and MCR, each spanning textual, prior-image, and visual-overlay modalities. Figure~\ref{fig:construction} summarizes how this case-level formulation is instantiated through source-case filtering, radiologist review, and context construction.

\paragraph{\textbf{IOR}, Image-only Recognition.}
The IOR family asks whether the model identifies the target finding from the current image alone. IOR is the no-context baseline, in which the model receives $I_i$ and the fixed 12-label option list. The hidden target $X_i$ is used only for scoring. IOR accuracy measures whether $\hat{p}_{i,\mathrm{ior}}$ matches $X_i$. Switch analyses condition on each model's IOR-correct subset to isolate context-induced changes from recognition error.

\paragraph{\textbf{RCS}, Reliable Context Stability.}
The RCS family asks whether the model remains stable when context agrees with the image. RCS measures whether reliable context preserves an image-only correct decision. Reliable contexts span all three modalities. The textual reliable contexts are the original clinical indication, the same-patient prior report, and a current-image-consistent preliminary note. The prior-image reliable context is the same-patient prior CXR. The visual-overlay reliable context is a pathology-aligned PACS-style arrow. RCS thus separates whether context agrees with the image from whether the model uses that agreement without losing its grip on the image evidence. The two are not the same. Reliable context still flips an image-only-correct prediction at non-trivial rates.

\paragraph{\textbf{MCR}, Misleading Context Resistance.}
The MCR family asks whether the model resists plausible but misleading context. MCR measures whether models resist context that plausibly implies $Y \neq X$ across all three modalities. The textual misleading contexts are a rewritten indication, prior report, or preliminary note. The prior-image misleading context is a cross-patient prior CXR. The visual-overlay misleading context is an off-pathology PACS-style arrow. Each case measures both whether the target-finding judgment flips and whether the predicted finding aligns with $Y$. MCR therefore measures conditional use. A model that uniformly trusts context fails MCR. A model that uniformly ignores context may remain stable, but it does not demonstrate beneficial use of reliable context.

\begin{table*}[t]
\centering
\footnotesize
\setlength{\tabcolsep}{3pt}
\begin{tabular*}{\textwidth}{@{\extracolsep{\fill}} l cc ccc cc ccc cc @{}}
\toprule
& \multicolumn{2}{c}{Image-only} & \multicolumn{5}{c}{Reliable context} & \multicolumn{5}{c}{Misleading context} \\
\cmidrule(lr){2-3} \cmidrule(lr){4-8} \cmidrule(lr){9-13}
& & & \multicolumn{3}{c}{Textual} & \multicolumn{2}{c}{Visual} & \multicolumn{3}{c}{Textual} & \multicolumn{2}{c}{Visual} \\
\cmidrule(lr){4-6} \cmidrule(lr){7-8} \cmidrule(lr){9-11} \cmidrule(lr){12-13}
Model & IOR & $\kappa$ & ind & rep & note & img & overlay & ind & rep & note & img & overlay \\
\midrule
InternVL3-8B           & 26.7 & 0.165 & 26.2 & 27.9 & \textbf{70.8} & 23.8 & 22.7 & 12.9 & 11.2 &  6.2 & 22.9 & 10.6 \\
Qwen3.5-9B             & 30.0 & 0.198 & \underline{37.1} & 37.9 & \underline{68.8} & \underline{32.9} & \textbf{28.8} & 25.0 & 17.5 &  9.2 & \textbf{30.8} & \underline{23.6} \\
Gemma-3-12B-IT         &  9.2 & 0.001 & 20.8 & 25.8 & \underline{68.8} &  8.8 & 15.3 &  5.8 &  3.8 &  4.6 & 10.0 &  6.0 \\
Llama-3.2-11B-V        &  8.3 & 0.008 & 22.9 & 28.3 & 58.3 &  7.5 & 15.3 &  5.4 &  4.6 &  3.3 &  8.8 & 11.1 \\
Phi-4-multimodal       & 15.8 & 0.066 & 23.8 & 27.5 & 66.2 & 12.1 & 11.0 &  7.9 &  7.1 &  4.2 &  9.2 &  7.5 \\
\midrule
MedGemma-1.5-4B        & \underline{31.7} & \underline{0.249} & 29.6 & 34.6 & 64.6 & 32.5 & \textbf{28.8} & 17.9 & 13.3 & \underline{12.1} & 16.2 & 19.6 \\
NV-Reason-CXR-3B       & \textbf{35.8} & \textbf{0.282} & \textbf{38.8} & \underline{39.6} & 58.8 & \textbf{36.7} & 21.5 & \textbf{33.3} & \textbf{27.1} & \textbf{20.4} & 27.1 & \textbf{32.2} \\
\midrule
Claude-Opus-4-7        & 21.2 & 0.115 & 31.7 & 35.0 & 66.2 & 14.2 & 20.9 & 20.4 & 11.2 & 11.7 & 16.7 & 10.6 \\
GPT-5.5                & 30.4 & 0.230 & \underline{37.1} & \textbf{41.2} & \textbf{70.8} & 30.8 & \textbf{28.8} & \underline{27.9} & 15.4 & 10.8 & \underline{29.2} & 13.6 \\
Gemini-3.5-Flash       & 28.3 & 0.184 & 30.8 & 34.2 & \underline{68.8} & 28.7 & \underline{27.0} & 25.8 & \underline{21.7} & 10.4 & 22.9 & 11.6 \\
\bottomrule
\end{tabular*}
\caption{Absolute accuracy across context conditions for the ten evaluated VLMs, together with image-only recognition (IOR) accuracy and Cohen's $\kappa$ as model-level summaries. These columns are evaluated without IOR-correct conditioning and therefore reflect conventional target-label classification accuracy. The high current-report-derived preliminary-note column reflects by-construction target evidence, whereas the low misleading-context columns reflect context-induced disruption under misleading conditions. Bold and underline mark the best and second-best result per column, respectively (higher is better).}
\label{tab:per_arm_acc}
\end{table*}

\subsection{Design Principles and Metrics}
\label{sec:design_metrics}

\paragraph{Design principles.}
Three principles guide construction. Under current-image priority, the benchmark is scored against the adjudicated current-image finding, while auxiliary context is varied only to test stability and resistance. Under paired perturbation, reliable and misleading conditions keep $I_i$ and $X_i$ fixed where both are defined. Under directional error attribution, misleading contexts imply a specific $Y$ so that $Y$-aligned errors can be attributed to context pull rather than random instability. The paired-perturbation principle follows behavioral and contrastive testing in NLP \citep{ribeiro2020checklist,gardner2020contrastsets}.

\paragraph{Failure-mode metrics.}
Two case-internal metrics define \mccxr{}'s failure-mode evaluation. Let $\mathcal{I}_M = \{i : \hat{p}_{i,\mathrm{ior}} = X_i\}$ denote the IOR-correct subset for model $M$, where $\mathrm{ior}$ is the image-only condition. The switch-to-wrong rate is the fraction of $\mathcal{I}_M$ whose prediction flips under context condition $c$.

\begin{equation}
P(C \rightarrow W \mid c) =
\frac{\#\{i \in \mathcal{I}_M : \hat{p}_{i,c} \neq X_i\}}
{|\mathcal{I}_M|}.
\end{equation}

For misleading-context conditions, we additionally compute the context-aligned error rate, the proportion of those switched errors whose predicted finding matches the context-implied label $Y$. Since $Y_i \neq X_i$ by construction, $\hat{p}_{i,c}=Y_i$ implies a switch.

\begin{equation}
P(W_Y \mid C \rightarrow W, c) =
\frac{\#\{i \in \mathcal{I}_M : \hat{p}_{i,c} = Y_i\}}
{\#\{i \in \mathcal{I}_M : \hat{p}_{i,c} \neq X_i\}}.
\end{equation}

$P(W_Y \mid C \rightarrow W, c)$ is undefined when a model has no switched errors under condition $c$ and is reported as NA in that case.

Switch-to-wrong measures flips in the target judgment, and context-aligned error checks whether those flips land on the specific class implied by the misleading context. Together they distinguish random instability from directional context pull \citep{ribeiro2020checklist,gardner2020contrastsets}. Raw target-finding accuracy is reported in Table~\ref{tab:per_arm_acc}. The constrained-letter protocol forces a choice from $\{$A,\ldots,L$\}$ so the model is offered no explicit abstention option, and responses that fail to yield a valid letter are marked invalid and counted as incorrect.

\subsection{Dataset Construction}
\label{sec:construction}

\mccxr{} is constructed from MIMIC-CXR \citep{johnson2019mimiccxr,johnson2019mimiccxrjpg} through a radiologist-gated pipeline. Eligible cases require a frontal PA or AP current CXR with the patient's most recent prior frontal study and parseable Findings or Impression sections in both reports. CheXpert source labels must correspond to a single confident positive abnormality or to No Finding, with uncertain ($-1.0$) and multi-positive cases excluded \citep{irvin2019chexpert}. The 12-label output space excludes Support Devices (non-diagnostic hardware) and Pleural Other (clinically ambiguous and under-represented in this cohort). Construction details are provided in Appendix~\ref{app:details}.

For each eligible case, the radiologist confirms a target finding $X$ on the current image and selects a single misleading label $Y$ that is clinically plausible, distinct from $X$, and not co-occurring with $X$ in the current report. The radiologist overrode the CheXpert source label for $X$ in 26 of the 240 cases (10.8\%), confirming non-trivial source-label noise and substantiating the radiologist-gated design. The same $Y$ is used across all misleading-context conditions for a case so that downstream context-aligned errors point in a known direction.

The radiologist then authors or verifies reliable and misleading conditions across five context sources spanning textual, prior-image, and visual-overlay modalities, under a side-by-side review interface. Misleading textual conditions rewrite the source text from $X$ toward $Y$ while preserving non-target clinical content, including anatomic location, laterality, severity, temporal wording, and negated findings. The cross-patient prior-image condition draws from priors whose CheXpert labels include $Y$ and exclude $X$, view- and demographics-matched by a deterministic ranking with the radiologist's final approval. The on-finding overlay condition is restricted to positive-abnormality cases because an on-pathology arrow is undefined for No Finding.

The current-report-derived preliminary-note condition is finalized from the current report and serves as a reference condition for the matched rewritten preliminary-note condition. The resulting benchmark contains 240 cases from 233 unique patients, expanded across 11 context conditions into 2{,}522 model-evaluable instances.

\begin{table}[t]
\centering
\footnotesize
\setlength{\tabcolsep}{4pt}
\begin{tabular}{l ccc cc}
\toprule
& \multicolumn{3}{c}{Textual} & \multicolumn{2}{c}{Visual} \\
\cmidrule(lr){2-4} \cmidrule(lr){5-6}
Model & ind & rep & note & img & overlay \\
\midrule
InternVL3-8B           & 63.9 & 67.4 & 83.9 & 20.0 & 22.5 \\
Qwen3.5-9B             & 73.7 & 64.5 & 93.0 & 33.3 & 20.0 \\
Gemma-3-12B-IT         & 57.9 & 65.0 & 86.4 & \underline{9.1} & \underline{5.6} \\
Llama-3.2-11B-V        & 85.7 & 68.8 & \underline{80.0} & 20.0 & 22.2 \\
Phi-4-multimodal       & 70.0 & \underline{58.6} & \textbf{79.4} & \textbf{4.2} & 16.0 \\
\midrule
MedGemma-1.5-4B        & 70.0 & 65.3 & 83.9 & 13.6 & 20.0 \\
NV-Reason-CXR-3B       & 72.7 & 64.3 & 85.7 & 14.3 & \textbf{0.0} \\
\midrule
Claude-Opus-4-7        & \underline{43.8} & 69.0 & 84.2 & 10.5 & 17.1 \\
GPT-5.5                & 70.6 & 74.4 & 87.5 & 26.7 & 22.0 \\
Gemini-3.5-Flash       & \textbf{33.3} & \textbf{50.0} & 88.5 & 28.6 & 20.0 \\
\bottomrule
\end{tabular}
\caption{Directional $Y$-alignment among switches induced by misleading context across the ten evaluated VLMs. Results report the fraction of switched errors that land on the context-implied label $Y$ for each misleading source. Textual conditions show systematic directional pull, whereas visual conditions produce lower and more scattered alignment. Bold and underline mark the best and second-best result per column, respectively, and lower values are better.}
\label{tab:per_arm_aligned}
\end{table}

\begin{figure*}[t]
\centering
\includegraphics[width=\linewidth]{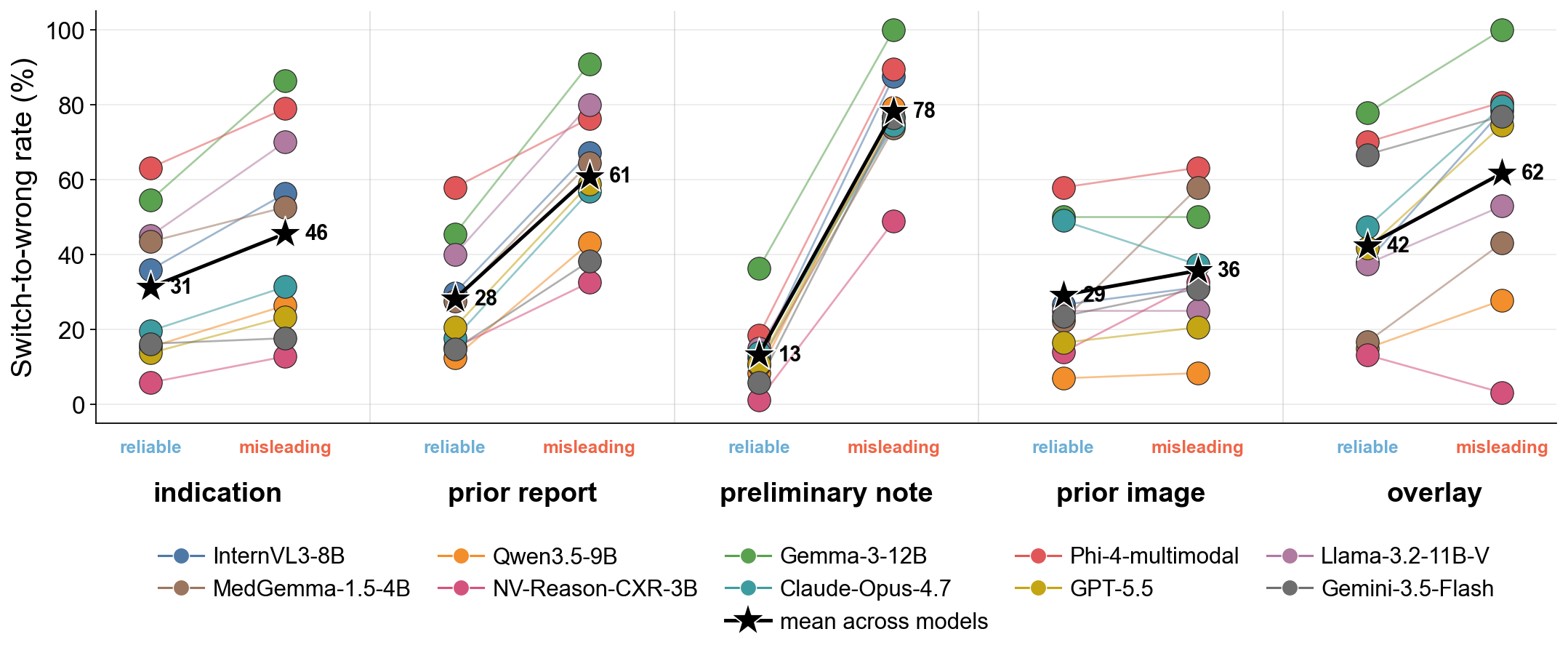}
\caption{Switch-to-wrong rate across context conditions on the IOR-correct subset for the ten evaluated VLMs (lower is better). Coloured dots are per-model rates and black stars mark the mean across models. The rewritten preliminary-note condition produces the strongest misleading-text disruption, the current-report-derived preliminary-note condition is the most stability-preserving reliable context, and the off-finding overlay condition shows strong visual perturbation without corresponding directional alignment.}
\label{fig:per_arm_switch}
\end{figure*}

\section{Experiments}
\label{sec:results}

\subsection{Evaluation Protocol}

Having formalized \mccxr{} as paired context perturbations, we evaluate whether current VLMs preserve image-grounded decisions across those matched conditions.

We evaluate ten VLMs in three categories. The first category comprises five general-purpose open-source models, namely InternVL3-8B \citep{chen2025internvl3}, Qwen3.5-9B \citep{qwen2026qwen35}, Gemma-3-12B-IT \citep{gemmateam2025gemma3}, Llama-3.2-11B-Vision-Instruct \citep{grattafiori2024llama3}, and Phi-4-multimodal-instruct \citep{kim2025phi4multimodal}. The second category comprises two CXR or medical-domain open-source models, namely MedGemma-1.5-4B \citep{sellergren2026medgemma15} and NV-Reason-CXR-3B \citep{myronenko2025nvreason}. The third category comprises three closed-source frontier models, namely Claude Opus 4.7 \citep{anthropic2025claude4} accessed via the Anthropic API, GPT-5.5 \citep{openai2026gpt55} accessed via the OpenAI API, and Gemini 3.5 Flash \citep{google2026gemini35flash} accessed via the Google AI Studio API. The primary protocol is zero-shot constrained-letter multiple-choice prompting. The model receives the image and condition context with a fixed A--L option list mapping to the 12 valid labels and emits a single capital letter.

Outputs are parsed by a single rule-based extractor, and responses that fail to yield a valid label are marked invalid and treated as incorrect. The main evaluation is performed on each model's IOR-correct subset. We first identify the cases that the model answers correctly under image-only input, and then measure whether the response switches under each context condition. This conditioning prevents baseline recognition failures from being counted as context-induced disruption. The size of this subset is determined by each model's IOR accuracy on the 240-case benchmark. It ranges from 20 cases for Llama-3.2-11B-V to 86 for NV-Reason-CXR-3B, and is smaller still for the overlay conditions, which draw on the retained core subset and, for on-finding, positive cases only. All reported numbers reflect a single deterministic run per model. Open-source models use greedy decoding and closed-source models are queried with deterministic settings, so we do not report run-to-run error bars. Full model identifiers, decoding settings, and compute are given in Appendix~\ref{app:compute}.

\subsection{Aggregate Results}

Table~\ref{tab:per_arm_acc} reports absolute accuracy for each context condition without IOR-correct conditioning. Cohen's $\kappa$ \citep{cohen1960kappa,landis1977measurement} between IOR predictions and gold $X$ peaks at 0.282 (NV-Reason-CXR-3B), with most models in the $0.00$--$0.25$ range. This range is conventionally interpreted as slight to fair agreement. Current VLMs are far from reliable single-image CXR classifiers even before any context is introduced. Two patterns stand out. The current-report-derived preliminary-note condition reaches a 66.2\% across-model mean because it provides target evidence by construction. All five MCR conditions collapse to across-model means below 20\%. The collapse is most severe for misleading preliminary notes (9.3\%) and mildest for cross-patient prior images (19.4\%). Per category, IOR across-model means are 18.0\% (open-source general), 33.8\% (medical-domain), and 26.7\% (closed-source). Under misleading preliminary notes, these drop to 5.5\%, 16.3\%, and 11.0\% respectively. The resulting image-only to rewritten preliminary-note gap of $12.5$--$17.5$ percentage points is largest for the medical-domain category, indicating that domain specialization does not insulate against textual override. Absolute accuracy alone, however, does not separate context-induced disruption from baseline recognition failure, motivating the switch and directional metrics that follow.

\section{Failure Patterns in Context Use}
\label{sec:failure}

Switch-based metrics conditioned on the IOR-correct subset isolate context-induced disruption from baseline recognition error. The three failure families below converge on a single asymmetry. Misleading textual context flips IOR-correct decisions and aligns them with the implied label, while misleading visual context flips them without directional commitment. This pattern connects CXR evaluation to shortcut learning \citep{geirhos2020shortcut}, sycophancy \citep{sharma2024sycophancy}, and knowledge conflict in retrieval-augmented LLMs \citep{xie2024chameleon,wang2025conflictingrag}. Figure~\ref{fig:per_arm_switch} and Table~\ref{tab:per_arm_aligned} provide the aggregate switch and directional-alignment views.

\subsection{Instability under Reliable Context}

In Reliable Context Stability, the context is not misleading by construction, yet models switch away from correct image-only answers at an across-model average of 28.8\% over the five conditions. This is an unweighted average because the on-finding overlay condition is restricted to positive-abnormality cases, for which on-pathology arrows can be placed. The per-source ordering is informative. The on-finding overlay (42.4\%) and reliable-indication (31.3\%) conditions are the most disruptive, the same-patient prior-image (29.2\%) and same-patient prior-report (28.1\%) conditions sit close to the across-model mean, and the concise current-report-derived preliminary-note condition (13.3\%) is the most stability-preserving. This exploratory ordering is consistent with sensitivity to differences in source register and visual salience, but the experiment does not isolate those factors. Even when the arrow points to the true pathology, models switch at non-trivial rates.

\subsection{Textual Context Over-Reliance}

The first headline of \mccxr{} is that VLMs are more directionally affected by misleading text than by misleading visual context. Under rewritten preliminary notes, averaged across models, 78.1\% of image-only-correct decisions flip and 85.3\% land on the context-implied label. Rewritten prior reports and clinical indications produce weaker but still systematic overrides, with flip rates of 60.9\% and 45.6\% and $Y$-alignment of 64.7\% and 64.2\%, respectively (Table~\ref{tab:per_arm_aligned}, Figure~\ref{fig:per_arm_switch}). This exploratory ordering is consistent with differences in source register, but authority, formality, and assertiveness were not independently manipulated. The note condition reaches up to 93.0\% $Y$-alignment on Qwen3.5-9B, providing model-level evidence of directed rather than diffuse errors in that condition. Comparable text-only knowledge-conflict and sycophancy studies report $40$--$80$\% override rates \citep{xie2024chameleon,sharma2024sycophancy,wang2025conflictingrag}.

\subsection{Visual Confusion vs.\ Language-Prior Bias}

The second headline of \mccxr{} mirrors the first. Misleading visual context flips predictions but does not direct them toward the misleading label. The cross-patient prior-image condition switches at 35.7\% with only 18.0\% $Y$-alignment, and the off-finding overlay condition switches at 61.7\% with only 16.5\% $Y$-alignment (Table~\ref{tab:per_arm_aligned}). These are high switch rates with scattered errors. Even the more aggressive visual perturbation does not reproduce the 85.3\% $Y$-alignment seen for rewritten preliminary notes, and the gap holds across all ten evaluated models under the same case-internal protocol. Pooled over the switched cases of all ten models, textual misleading context produces 713/956 (74.6\%) $Y$-aligned errors against 78/443 (17.6\%) for visual context, a 57.0-percentage-point gap (case-cluster-bootstrap 95\% CI 50.9--62.8); every model individually shows the same direction (two-sided sign test $p=0.002$). This asymmetry is consistent with stronger directional pull from textual context under the standardized direct-answer protocol.

The directional gap remains large after excluding the strongest misleading-text condition, the rewritten preliminary note: 345/528 (65.3\%) for text versus 78/443 (17.6\%) for visual context, a 47.7-point gap (case-cluster-bootstrap 95\% CI 40.9--54.3). Descriptive threshold analyses also yield similar gaps when retaining models with IOR accuracy at least 20\% (55.9 points) or $\kappa$ at least 0.15 (55.6 points), although these checks cannot remove chance-correct cases from the conditioned subsets.

Across these analyses, models do not always preserve their image-only-correct decisions when misleading context is present under the standardized direct-answer protocol.

\section{Conclusion}

\mccxr{} measures whether vision--language models preserve image-grounded decisions when clinical context is added. Across ten models and five context sources, the answer is asymmetric. VLMs abandon correct image-only decisions under misleading text, with directional alignment reaching 85\% for rewritten preliminary notes, while visually misleading cues produce scattered predictions. This asymmetry is observed under the standardized direct-answer protocol and is related to sycophancy and knowledge conflict in text-only LLMs. We release \mccxr{} as a paired benchmark for evaluating context robustness and corresponding interventions in medical VLMs.

\section*{Limitations}

Five caveats bound \mccxr{}'s findings. First, IOR accuracy is 8.3--35.8\% and Cohen's $\kappa$ against gold $X$ peaks at 0.282, so IOR-correct conditioning isolates context-induced change but does not rule out chance-correct cases. Second, the numbers reflect single deterministic runs under a constrained-letter direct-answer prompt. A GPT-5.5 conflict-aware pilot (Appendix~\ref{app:diagnostics}) reduces misleading-text switches, so the pattern is protocol-dependent rather than prompt-invariant. Third, curation is single-rater \citep{landis1977measurement}. The 51-case second-reader audit (Appendix~\ref{app:diagnostics}) covers b2/d2/e2, supports directional $Y$-implication, and does not establish $\kappa$ or benchmark-wide validation. Except for two minimally reconstructed indications, the reliable indications, prior reports, and prior CXRs are authentic source materials rather than validated to imply exactly $X$, so RCS switches under reliable context can reflect partial contextual mismatch in addition to instability under agreeing evidence. Fourth, evaluated models' MIMIC-CXR overlap cannot be audited; \mccxr{} makes no contamination-free claim. Fifth, the 12-label constrained-letter output offers no abstention and is not equivalent to free-text reporting. The current-report-derived preliminary-note condition (66.2\%, Table~\ref{tab:per_arm_acc}) is a reference control, and on-/off-finding overlays are evaluated on the retained core subset only. The misleading visual conditions (cross-patient prior CXR, off-finding overlay) were radiologist-curated to imply $Y$ but not source-only calibrated for $Y$ recognizability, so the low visual $Y$-alignment may partly reflect weaker cue specificity in addition to modality-preference asymmetry. Multi-rater curation, richer decoding, abstention, and larger cohorts are planned.

\section*{Ethical Considerations}

\mccxr{} is intended solely for research on multimodal model evaluation and should not support autonomous clinical interpretation \citep{degrave2021ai}. Annotation and curation were performed by a single board-certified radiologist who is an author of this work. No external annotators were recruited and no compensation applies (criteria in Appendix~\ref{app:validation}). Original MIMIC-CXR radiographs are not redistributed, and access to the underlying MIMIC-CXR data remains subject to the \href{https://physionet.org/content/mimic-cxr/view-license/1.0.0/}{PhysioNet Credentialed Health Data License}.

The benchmark's intentionally misleading clinical context is necessary for robustness evaluation and should be reported only in that setting; models evaluated on \mccxr{} should not be described as clinically safe or unsafe from benchmark performance alone.

\section*{Acknowledgments}

This work was supported by the National Research Foundation of Korea (NRF) grant funded by the Korea government (MSIT) (No.\ RS-2026-25479661; K.S.C.), the SNUH Research Fund (No.\ 04-2025-2060; K.S.C.), the Korea Health Technology R\&D Project through the Korea Health Industry Development Institute (KHIDI) funded by the Ministry of Health and Welfare (No.\ RS-2024-00439549; K.S.C.), the ``Advanced GPU Utilization Support Program'' funded by the Government of the Republic of Korea (Ministry of Science and ICT), and partly by the Institute of Information \& Communications Technology Planning \& Evaluation (IITP) through the AI Computing Support Project for R\&D funded by the Korea government (MSIT) (No.\ RS-2026-25505492, ``High-Performance Research AI Computing Infrastructure Support at the 2\,PFLOPS Scale'').

\bibliography{mc_cxr_references}

\clearpage

\appendix

\section{Details of MC-CXR Construction}
\label{app:details}

\subsection{Source Data and Filtering}

\mccxr{} is constructed from MIMIC-CXR \citep{johnson2019mimiccxr} and MIMIC-CXR-JPG \citep{johnson2019mimiccxrjpg}. The 240-case EMNLP cohort is drawn from study-id-disjoint MIMIC-CXR splits with radiologist-verified labels. Cases with Pleural Other are excluded before evaluation. We require a frontal PA or AP current CXR with the patient's most recent prior frontal study and parseable Findings or Impression sections in both current and prior reports. CheXpert-style source labels must correspond to either a single confident positive abnormality or to No Finding. We exclude any case with uncertain ($-1.0$) labels or with multiple confident positives, and we remove the Support Devices label from the target pool. The core subset is stratified-sampled across the resulting label space to balance the X-label distribution. For each case, the reliable prior is drawn from the same patient, whereas the misleading prior is drawn from a different patient. Final benchmark files provide reproducible image identifiers rather than raw radiographs.

\subsection{Target and Misleading Label Pairing}

For each eligible image, the radiologist confirms an effective target finding $X$. The initial $X$ is taken from the source label table but is overridden if judged incorrect by the radiologist, with 26 such overrides across the 240-case benchmark (10.8\%). The misleading label $Y$ is drawn from the same pool such that it is clinically plausible, distinct from $X$, and not co-occurring with $X$ in the current report. Cases in which $Y$ is uncertain or visually ambiguous are excluded. The same $Y$ is used across every misleading-context condition for a case.

\subsection{Context-Condition Construction}

The reliable-indication condition uses the original indication where available, with two missing indications minimally reconstructed; the reliable prior-report condition uses the same-patient prior report, and the reliable prior-image condition uses the same-patient prior CXR. The current-report-derived preliminary-note condition is finalized by the radiologist from the current report's findings and impression. The misleading indication, prior-report, and preliminary-note conditions are written by the radiologist under a side-by-side review interface, rewriting the source text from $X$ toward $Y$ while preserving non-target clinical content such as anatomic location, laterality, severity, temporal wording, and negated findings. The cross-patient prior-image condition uses prior CXRs from different patients whose CheXpert labels include $Y$ and exclude $X$, view- and demographics-matched by a deterministic ranking with the radiologist selecting the final candidate after image inspection. The on-finding and off-finding overlays are placed or verified directly by the radiologist over the current image. The on-finding overlay condition is restricted to positive-abnormality cases because an on-pathology arrow is undefined for No Finding.

\subsection{Label-Space Construction}

The evaluation uses a 12-label option list. The labels are No Finding, Enlarged Cardiomediastinum, Cardiomegaly, Lung Opacity, Lung Lesion, Edema, Consolidation, Pneumonia, Atelectasis, Pneumothorax, Pleural Effusion, and Fracture. Support Devices and Pleural Other are excluded from the gold pool. Support Devices is a non-diagnostic hardware finding and was never selected as $X$ or $Y$. Pleural Other is dropped because it is under-represented and clinically ambiguous in the curated cohort. The model is free to predict any of the 12 labels, allowing us to detect off-target context pull beyond $Y$-aligned errors.

\section{Question Templates}
\label{app:templates}

\subsection{Constrained-Letter Multiple-Choice Template}
\label{app:letter_mc_template}

\begin{quote}\small
You are a radiologist classifying a chest X-ray. Choose exactly one option from the lettered list below. Reply with a single capital letter only, with no other text.

[Image context block: one current image, or two images when prior context is provided]

[Optional context: INDICATION / PRIOR REPORT / PRELIM NOTE]

Choose the most likely finding:
A. Pneumonia
B. Edema
C. Atelectasis
D. Consolidation
E. Cardiomegaly
F. Pleural Effusion
G. Pneumothorax
H. Lung Opacity
I. Lung Lesion
J. Enlarged Cardiomediastinum
K. Fracture
L. No Finding

Answer with a single letter (A-L):
\end{quote}

\section{Evaluation Configuration and Compute}
\label{app:compute}

\paragraph{Model access.} Open-source models are served locally with the HuggingFace \texttt{transformers} library (\texttt{trust\_remote\_code} enabled): InternVL3-8B (\texttt{OpenGVLab/InternVL3-8B-hf}), Qwen3.5-9B (\texttt{Qwen/Qwen3.5-9B}), Gemma-3-12B-IT (\texttt{google/gemma-3-12b-it}), Llama-3.2-11B-Vision-Instruct (\texttt{meta-llama/Llama-3.2-11B-Vision-Instruct}), Phi-4-multimodal-instruct (\texttt{microsoft/Phi-4-multimodal-instruct}), MedGemma-1.5-4B (\texttt{google/medgemma-1.5-4b-it}), and NV-Reason-CXR-3B (\texttt{nvidia/NV-Reason-CXR-3B}). Closed-source models are accessed through their APIs in May 2026 as \texttt{anthropic/claude-opus-4-7}, \texttt{openai/gpt-5.5}, and \texttt{gemini/gemini-3.5-flash}; their parameter counts are not publicly disclosed.

\paragraph{Decoding.} Open-source models use deterministic greedy decoding (\texttt{do\_sample}=false, \texttt{bfloat16} weights, \texttt{device\_map}=auto) with a budget of \texttt{max\_new\_tokens}=8, sufficient for the single-letter answer. Closed-source models are queried through a unified API layer at \texttt{temperature}=0 with \texttt{max\_tokens}=1024 (to accommodate internal reasoning tokens) and minimal reasoning effort. Image preprocessing uses each model's native processor at its default resolution caps.

\paragraph{Compute.} Open-source models run on a single node with 4$\times$NVIDIA RTX A6000 (49\,GB each). Each model evaluates all 2{,}522 instances in roughly 12--125 minutes depending on architecture, so the full open-source sweep takes on the order of ten GPU-hours. Closed-source models are evaluated through their APIs and incur API cost only.

\section{Response-Period Diagnostics and API Provenance}
\label{app:diagnostics}

\subsection{Second-Reader Audit Aggregate}

A stratified 51-case subset audit was reviewed by an independent board-certified reader blinded to model outputs, paper results, and first-reader case-level ratings. Sampling targeted four cases per $X$-label class where available, with sparse-class shortfalls redistributed and remaining slots filled across strata to reach 51. The audit covers only the three misleading textual conditions (b2, d2, e2). Confirmation, equivocal, and concern counts (of 51 per row) are as follows. $X$ valid: 43 / 6 / 2. $Y$ plausible: 31 / 16 / 4. b2 implies $Y$: 51 / 0 / 0. d2 implies $Y$: 51 / 0 / 0. e2 implies $Y$: 51 / 0 / 0. b2 has no extra non-$Y$ finding: 44 / 6 / 1. d2 has no extra non-$Y$ finding: 37 / 9 / 5. e2 has no extra non-$Y$ finding: 51 / 0 / 0. Equivocals are retained separately from confirmations and concerns. Cohen's $\kappa$ is not computed because the first reader did not provide paired ratings on the same categorical audit scale. This audit describes the sampled subset; it does not establish benchmark-wide inter-rater validity, prior-image or overlay validation, or a confirmed-only performance analysis.

\subsection{Prompt Ablation Pilot (GPT-5.5)}

To probe whether the misleading-text switch behavior is prompt-invariant, we ran a fixed-original-cohort sensitivity on the 73 cases correct in the paper's GPT-5.5 image-only run. Three system instructions were compared on the misleading indication (b2), prior report (d2), and preliminary note (e2) conditions: PB, the paper's baseline direct-answer instruction; PI, an image-first instruction that names the current image as the primary evidence; and PC, a conflict-aware instruction that instructs the model to prefer the image when context disagrees. Only the system instruction varied. Contemporaneous PB switch rates were d2 40/73 (54.8\%) and e2 56/73 (76.7\%). PI showed no clear paired change (d2 57.5\%, e2 75.3\%). PC reduced switches to d2 27/73 (37.0\%) and e2 30/73 (41.1\%), a paired McNemar reduction of $-17.8$~pp (95\% CI $[-28.8, -8.2]$, $p=.0023$) and $-35.6$~pp (95\% CI $[-46.6, -24.7]$, $p<.0001$) respectively. This is a fixed-original-cohort prompt sensitivity, not a contemporaneous switch-rate re-estimation: matched reliable and visual conditions were not rerun, the closed-API alias was not version-pinned (5 d2 and 8 e2 predictions differ between the paper and the contemporaneous PB runs), and only one model and two textual conditions were evaluated. The reduction indicates that a conflict-aware prompt can attenuate misleading-text override, but it does not establish improved conditional evidence use, closure of the text--visual gap, or invariance across models, prompts, or token budgets.

\subsection{Closed-API Provenance}

Closed-source models were queried in May 2026 through provider service aliases rather than version-pinned snapshots. Table~\ref{tab:api_provenance} lists the alias, provider SDK, decoding parameters at query time, and pinning availability. The Anthropic, OpenAI, and Google APIs did not return an immutable backend snapshot identifier in the response envelope; the aliases are therefore the only stable handle. Exact closed-API backend stability is not assumed, and the prompt pilot uses the contemporaneous PB run rather than the paper's stored predictions when computing intervention deltas.

\begin{table}[t]
\centering
\scriptsize
\setlength{\tabcolsep}{3pt}
\begin{tabular}{@{}llll@{}}
\toprule
Model & Alias & SDK & Pinning \\
\midrule
Claude Opus 4.7 & \texttt{claude-opus-4-7} & \texttt{anthropic} & alias \\
GPT-5.5 & \texttt{gpt-5.5} & \texttt{openai} & alias \\
Gemini 3.5 Flash & \texttt{gemini-3.5-flash} & \texttt{google-genai} & alias \\
\bottomrule
\end{tabular}
\caption{Closed-source API provenance. All three providers were queried in May 2026 through service aliases; no immutable backend snapshot identifier was exposed in the response envelope. Decoding used temperature=0, max\_tokens=1024, minimal reasoning effort.}
\label{tab:api_provenance}
\end{table}

\section{Validation Protocol}
\label{app:validation}

Text contexts that would introduce unsupported findings beyond the controlled $X$-toward-$Y$ rewriting are revised or replaced. Prior-image misleading contexts are revised or replaced if the retrieved image does not visibly support $Y$ or if it also supports $X$. Overlay placements are verified for the 163 on-finding and 199 off-finding cases where the overlay is defined; the overlay conditions are not defined for the full 240-case cohort.

\end{document}